\documentclass{article}

\usepackage{PRIMEarxiv}

\usepackage[utf8]{inputenc} % allow utf-8 input
\usepackage[T1]{fontenc}    % use 8-bit T1 fonts
\usepackage{hyperref}       % hyperlinks
\usepackage{url}            % simple URL typesetting
\usepackage{booktabs}       % professional-quality tables
\usepackage{amsfonts}       % blackboard math symbols
\usepackage{nicefrac}       % compact symbols for 1/2, etc.
\usepackage{microtype}      % microtypography
\usepackage{lipsum}
\usepackage{fancyhdr}       % header
\usepackage{graphicx}       % graphics
\graphicspath{{media/}}     % organize your images and other figures under media/ folder

\title{Improving Direct Persian–English Speech-to-Speech Translation with Discrete Units
and Synthetic Parallel Data
}

\author{
  Sina Rashidi, Hossein Sameti \\
  Computer Engineering Department \\
  Sharif University of Technology \\
  Tehran, Iran\\
  \texttt{\{sina.rashidi, sameti\}@sharif.edu} \\
}

\begin{document}
\maketitle

\begin{abstract}
Direct speech-to-speech translation (S2ST), in which all components are trained jointly,
is an attractive alternative to cascaded systems because it offers a simpler pipeline and
lower inference latency. However, direct S2ST models require large amounts of parallel
speech data in the source and target languages, which are rarely available for low-resource
languages such as Persian. This paper presents a direct S2ST system for translating Persian speech into English speech, as well as a pipeline for synthetic parallel Persian-English speech generation.. 
% The model operates purely on speech, without relying on intermediate textual representations.
% Our approach combines a self-supervised pretrained speech encoder, a discrete-unit
% decoder, and a neural unit-based vocoder. 
The model comprises three components: (1) a conformer-based encoder,
initialized from self-supervised pretraining, maps source speech to high-level acoustic
representations; (2) a causal transformer decoder with relative position multi-head attention
translates these representations into discrete target speech units; (3) a unit-based neural vocoder generates waveforms from the predicted discrete units. To mitigate the data scarcity problem,
we construct a new Persian–English parallel speech corpus by translating Persian speech
transcriptions into English using a large language model and then synthesizing the
corresponding English speech with a state-of-the-art zero-shot text-to-speech system. The resulting corpus increases the amount of available parallel speech by roughly a factor of six.
On the Persian–English portion of the CVSS corpus, the proposed model achieves
improvement of  4.6 ASR BLEU with the
synthetic data over direct baselines. These results indicate that combining self-supervised
pretraining, discrete speech units, and synthetic parallel data is effective for
improving direct S2ST in low-resource language pairs such as Persian–English. \footnote{Code is publicly available at \href{https://github.com/sinarashidi/S2ST-Transformer}{https://github.com/sinarashidi/S2ST-Transformer}}
\end{abstract}

% keywords can be removed

\section{Introduction}
Speech-to-speech translation (S2ST) systems facilitate cross-lingual communication by converting speech in a source language directly into synthesized speech in a target language. Traditional S2ST pipelines decompose this task into three independent
modules: automatic speech recognition (ASR), text-based machine translation (MT),
and text-to-speech (TTS) synthesis \cite{wahlster2013verbmobil}. While modular systems have been highly successful,
they suffer from several drawbacks such as error propagation across modules, increased latency due to multiple processing stages, and multiplied training data requirements across subsystems.

\begin{figure}
  \centering
  \includegraphics[width=0.6\linewidth]{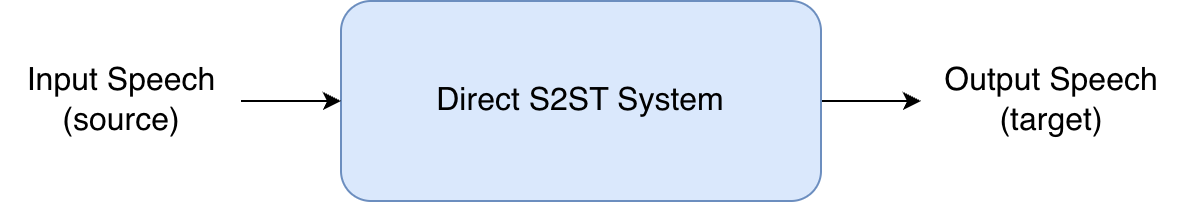}
  \caption{Demonstration of a direct speech-to-speech translation pipeline}
  \label{fig:direct_s2st}
\end{figure}

Direct, or end-to-end, S2ST models (\autoref{fig:direct_s2st}) address these issues by learning a single model that
maps source speech to target speech. By avoiding intermediate text representations,
direct systems can reduce inference time and simplify deployment \cite{jia2019direct}, which is particularly
appealing for audio dubbing scenarios where many utterances must be processed with
limited delay. However, direct S2ST models require large amounts of parallel speech
data, and existing corpora are heavily skewed toward a small set of high-resource
languages such as English and Spanish \cite{jia2019direct, jia2022translatotron, lee2021direct}.

For low-resource languages like Persian, collecting large volumes of parallel speech is
expensive and time-consuming. In addition, S2ST models for audio dubbing must not
only translate content accurately but also generate natural-sounding speech in the target
language. This combination of translation quality and speech naturalness makes data
scarcity an even more pressing challenge.

In this work we focus on direct Persian–English S2ST for audio dubbing. Building on
recent advances in discrete speech units, self-supervised speech representation learning,
and unit-based neural vocoders, we propose a model that translates Persian speech into
English speech without relying on intermediate text. Furthermore, we construct a new
synthetic Persian–English parallel speech corpus using large language models (LLMs) and neural
TTS, substantially expanding the amount of paired data available for this low-resource
language pair.

Our contributions are as follows:
\begin{enumerate}
    \item We present a direct S2ST architecture for Persian–English dubbing that combines a
pretrained conformer-based encoder, a causal transformer decoder with relative
positional attention, and a unit-based neural vocoder.
    \item We design a data generation pipeline that creates a large synthetic parallel speech
corpus by translating existing Persian speech transcriptions with a large language
model and synthesizing English speech using a zero-shot TTS system. The resulting
corpus increases the amount of available parallel speech data by about six times
compared to existing resources.
    \item We conduct a comprehensive experimental study on the CVSS Fa–En benchmark and
the newly constructed corpus, comparing our system to strong direct baselines such as
Translatotron and a speech-to-unit model.
    \item We analyze the impact of self-supervised pretraining, discrete units, and synthetic
data, showing that their combination yields consistent improvements in ASR BLEU
for Persian–English S2ST.
\end{enumerate}

\section{Background}
Direct S2ST has evolved rapidly over the past few years. Early works on speech translation
focused on cascaded systems in which ASR, MT, and TTS were trained independently \cite{wahlster2013verbmobil}.
More recently, sequence-to-sequence architectures have enabled direct speech-to-text
translation, and subsequently direct S2ST, by learning a unified mapping from source
speech to target output.

Spectrogram-based direct S2ST models such as Translatotron \cite{jia2019direct} and Translatotron 2 \cite{jia2022translatotron} use
an encoder–decoder architecture that predicts target mel-spectrograms, which are then
converted to waveform by a neural vocoder. These models have demonstrated that
direct S2ST can rival cascaded systems in high-resource settings and can preserve
speaker characteristics, which is valuable for dubbing applications. However, operating
entirely in the spectrogram domain can make training unstable and data-hungry, and
spectrogram prediction models may struggle to disentangle linguistic content from
speaker and prosodic information.

A complementary line of work replaces continuous spectrograms with discrete speech
units learned by self-supervised speech models followed by vector quantization or
clustering \cite{lee2021direct}. Discrete units serve as a compact, language-agnostic representation of speech,
enabling modular combinations of unit-based ASR, MT, and TTS components. For
S2ST, discrete units allow the translation network to focus on symbolic sequences, while
a separate unit vocoder handles waveform synthesis. Recent systems such as direct
S2ST with discrete units \cite{lee2021direct} and two-pass architectures like UnitY \cite{inaguma2022unity} demonstrate strong
performance by leveraging self-supervised pretraining and large-scale unit discovery.
Self-supervised representation learning for speech, exemplified by wav2vec 2.0 \cite{baevski2020wav2vec} and
related models, has proven highly effective in low-resource scenarios. By pretraining on
unlabeled speech and fine-tuning on task-specific data, these models substantially reduce
the amount of labeled data needed to achieve reliable performance. Data augmentation
techniques such as SpecAugment \cite{park2019specaugment} further improve robustness by applying time warping
and time/frequency masking to spectrograms during training.

Despite these advances, relatively little work has focused on low-resource language pairs
such as Persian–English, and even fewer on direct S2ST without intermediate text. This
paper aims to bridge that gap by combining self-supervised pretraining, discrete units,
and synthetic parallel data in a single direct S2ST system tailored to Persian–English
dubbing.

\section{Datasets}
Our experiments rely on three types of speech corpora and a synthetic Persian–English parallel corpus. We use the Persian portion of Common Voice \cite{ardila2019common}, a large crowdsourced corpus of read speech containing thousands of speakers and substantial hours of audio, both for self-supervised pretraining and as the starting point for building the synthetic parallel corpus. As our main benchmark for evaluating translation quality, we adopt the Persian–English subset of CVSS \cite{jia2022cvss}, a multilingual S2ST corpus constructed by aligning speech-to-text translation data and synthesizing target speech with a neural TTS system. In addition, for training and adapting the neural unit vocoder used in our pipeline, we employ LJSpeech \cite{ljspeech17}, an English TTS dataset consisting of high-quality recordings by a single speaker.

\subsection{Synthetic Persian-English Parallel Speech Corpus}
To alleviate the lack of Persian–English parallel speech, we construct a new synthetic
corpus in three stages.
\begin{enumerate}
    \item Collecting Persian speech and transcriptions.
We start from the Persian portion of Common Voice, which provides utterances along
with sentence-level transcriptions. After basic cleaning and filtering, we obtain a set of
high-quality Persian speech segments with reliable text.
    \item Translating Persian transcriptions to English with a large language model.
Each Persian transcription is translated into English using a state-of-the-art large
language model (GPT-4o). The model is prompted to produce fluent, semantically faithful
translations suitable for spoken dialogue. This yields synthetic English text that is
aligned at the utterance level with the original Persian speech.
    \item Synthesizing English speech with neural TTS.
The translated English sentences are converted into speech using a state-of-the-art zero-shot TTS model, VoiceCraft \cite{peng2024voicecraft}. The result is an English speech utterance aligned with the original
Persian speech at the utterance level. By repeating this process for all items, we obtain
a large synthetic Persian–English parallel speech corpus.
\end{enumerate}
Overall, the constructed corpus contains roughly six times more Persian–English parallel
speech than the existing CVSS Fa–En subset. This substantial increase in data volume
is crucial for training data-hungry direct S2ST models.

\section{Model}
\begin{figure}
  \centering
  \includegraphics[width=0.8\linewidth]{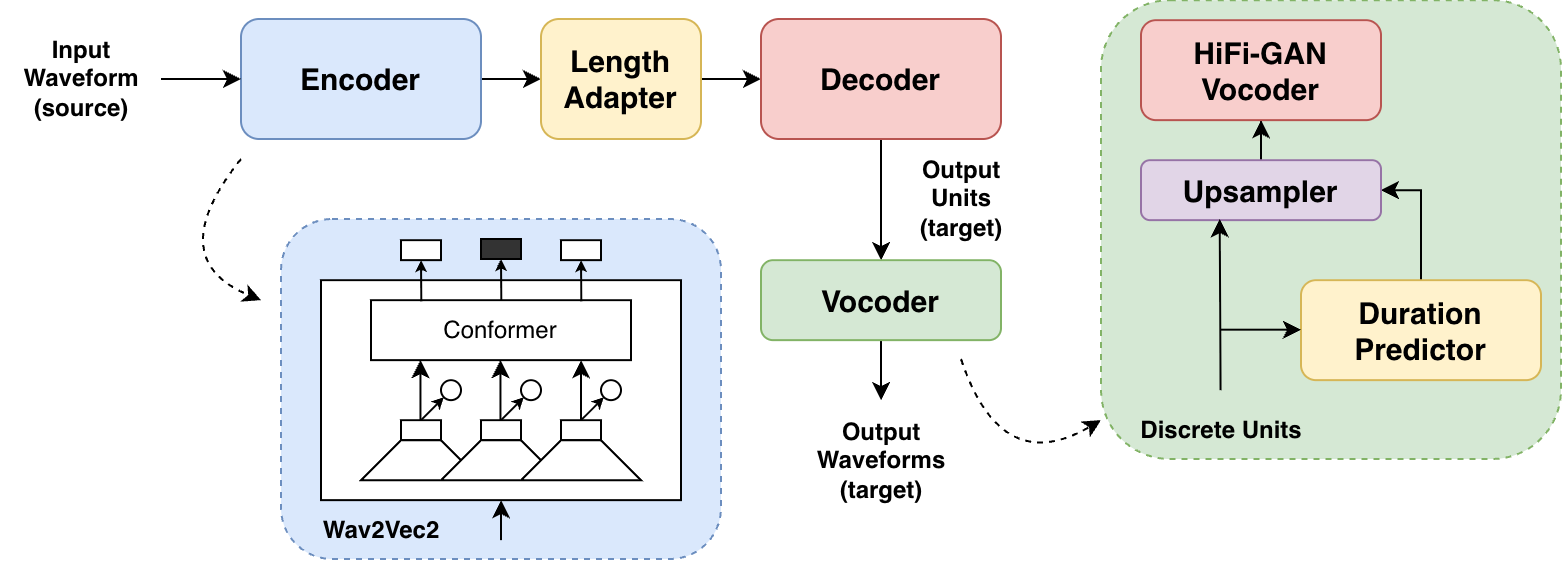}
  \caption{Architecture of the proposed model}
  \label{fig:model}
\end{figure}

The model follows a direct S2ST paradigm based on discrete speech units. It (\autoref{fig:model})
consists of three main components: (1) a conformer-based speech encoder \cite{gulati2020conformer} initialized from self-supervised pretraining on Persian speech; (2) a causal transformer decoder with relative position multi-head attention, which maps encoder representations to discrete target speech units; (3) a neural vocoder that converts sequences of discrete units into English speech waveforms.
The entire model is trained to maximize the likelihood of the target unit sequence given
the source speech, without accessing intermediate source or target text. The unit vocoder
is trained separately on English data and kept fixed during S2ST training.

\subsection{Speech Encoder}
The encoder is based on wav2vec 2.0, with its Transformer encoder layers replaced by Conformer layers. We first perform self-supervised pretraining on
the Persian portion of Common Voice using a contrastive objective similar to wav2vec
2.0. During pretraining, random spans of the input are masked, and the model learns to
distinguish the true latent representation of each masked region from a set of negative
examples. This encourages the encoder to capture robust, high-level acoustic features
that generalize well across tasks and domains.

After pretraining, the encoder is fine-tuned as part of the direct S2ST model. The
combination of self-supervised pretraining and supervised fine-tuning improves data
efficiency and robustness, especially in the low-resource Persian setting.

\subsection{Discrete Target Units and Decoder}
On the target side, we represent English speech using discrete units obtained from a
unit-based speech model trained on English data. Following prior work, we derive these
units by (i) learning frame-level latent representations with HuBERT \cite{hsu2021hubert}, a self-supervised speech
model and (ii) applying clustering to obtain a finite codebook of units. Each target
utterance is thus mapped to a sequence of discrete indices.

The S2ST decoder is a causal transformer that autoregressively predicts the next unit
given the previous units and the encoder outputs. We use relative position multi-head
attention \cite{shaw2018self} to better model the temporal structure of speech and to allow the decoder to
generalize to different utterance lengths. The decoder is trained with a cross-entropy
objective over unit indices, optionally combined with label smoothing.

\subsection{Length Adapter and Alignment}
Due to differences in speaking rate and phonotactics between Persian and English, the
lengths of source and target sequences may differ substantially. We incorporate a length
adapter that bridges the temporal resolution between encoder representations and target
units. The adapter uses simple convolutional and subsampling operations to reduce the
time dimension while preserving relevant information for translation. This module is
trained jointly with the encoder and decoder.

\subsection{Unit Vocoder}
The final component of the system is the neural unit vocoder that converts sequences of
discrete units into English speech waveforms. The vocoder is trained on English speech
and unit sequences derived from the LJSpeech corpus. In this work,
we adopt HiFi-GAN \cite{kong2020hifi} that has been shown to produce natural-sounding
speech from discrete units. Once trained, the vocoder is frozen and reused for all S2ST
experiments.

\subsection{Data Augmentation}
To improve robustness and reduce overfitting, we apply SpecAugment. Time shifting,
frequency masking, and time masking are randomly applied to the encoder inputs.
These perturbations encourage the model to focus on invariant acoustic cues and have
been shown to improve performance in both ASR and S2ST tasks.

\section{Experiments}
\subsection{Training Setup}
All speech signals are resampled to 16 kHz. We use dynamic batch construction based
on the total number of frames to efficiently utilize GPU memory. The main S2ST model
is trained for 40 epochs with an initial learning rate of 2.5e-4 for the decoder and a
smaller learning rate of 1e-5 for the pretrained encoder, using an optimizer with warmup
and decay. The training loss is computed as the average cross-entropy over all target
units in each mini-batch. 
We train two versions of the proposed model:
\begin{enumerate}
    \item CVSS-only: trained on the CVSS Fa–En subset.
    \item CVSS+Synthetic: trained on the combination of CVSS and the newly constructed
synthetic Persian–English corpus.
\end{enumerate}

\subsection{Baselines}
We compare our approach to several direct S2ST baselines:
(1) Translatotron: a spectrogram-based direct S2ST model that predicts target mel-
spectrograms from source speech and uses a neural vocoder to synthesize waveform;
(2) Translatotron 2 (+pretraining): an improved version with stronger pretraining and
architectural refinements;
(3) Speech-to-unit (+pretraining): a model that directly predicts discrete target units from
source speech but uses a simpler encoder–decoder architecture than our proposed
system.
These baselines represent strong direct S2ST systems and provide a meaningful point of
comparison for our contributions in encoder pretraining, unit modeling, and data
augmentation.

We evaluate translation quality using ASR BLEU. In this metric, the synthesized English
speech is fed to an English ASR model trained on LibriSpeech. The resulting
transcripts are compared against the reference English transcripts using the BLEU
metric. ASR BLEU correlates with translation quality while accounting for both
translation and synthesis errors.

We also report qualitative observations on speech naturalness and alignment based on
listening to model outputs, but the main quantitative comparisons are in terms of ASR
BLEU.

\subsection{Results}
\autoref{tab:results} summarizes the ASR BLEU scores on the CVSS Fa–En evaluation set for training with both CVSS-only and CVSS+Synthetic datasets. The proposed model achieves the highest ASR BLEU among all systems, outperforming
the best baseline (Translatotron 2 with pretraining) by 0.3 BLEU. This gain is achieved
despite using the same training data, indicating that the combination of a pretrained
conformer encoder, discrete target units, and a tailored decoder architecture provides a
more effective mapping from Persian speech to English speech units.

When we augment training with the synthetic Persian–English corpus, the proposed
model benefits substantially from the increased data. Relative to direct baselines trained
only on existing datasets, our system achieves gains of 0.3 ASR BLEU without synthetic
data and 4.6 ASR BLEU with synthetic data.
These improvements highlight the effectiveness of large-scale synthetic data for
low-resource S2ST, particularly when combined with self-supervised pretraining and
unit-based modeling.

Qualitatively, we observe that models trained with synthetic data produce more fluent
and semantically complete translations, especially for longer utterances and less frequent
phrases. The additional coverage of vocabulary and sentence patterns provided by the
synthetic corpus appears to reduce omissions and mistranslations in the generated
speech.

\begin{table}
  \caption{ASR BLEU (Fa → En)}
  \centering
  \begin{tabular}{lcc}
    \toprule
    Model & CVSS-Only & CVSS+Synthetic \\
    \midrule
    Translatotron & 1.4 & 6.9 \\
    \midrule
    Translatotron 2 & 2.4 & - \\
    + Pre-training & 3.8 & - \\
    \midrule
    Speech-to-unit & 1.6 & 11.8 \\
    + Pre-training & 2.8 & 13.2 \\
    \midrule
    Our Proposed model & \textbf{4.1} & \textbf{17.8} \\
    \bottomrule
  \end{tabular}
  \label{tab:results}
\end{table}

\section{Discussion and Conclusion}
In this work, we presented a direct Persian–English S2ST system designed for audio dubbing that integrates a self-supervised pretrained Conformer encoder, a discrete-unit Transformer decoder with relative positional attention, and a neural unit-based vocoder. To mitigate the scarcity of parallel data, we constructed a synthetic Persian–English corpus using large language model translation and neural TTS synthesis, increasing the amount of parallel speech by approximately six times. On the CVSS Fa–En benchmark, our model achieves up to a 4.6 BLEU improvement over strong direct baselines when trained with the synthetic corpus. These results demonstrate that the combination of self-supervised pretraining, discrete units, and synthetic data is an effective strategy for improving S2ST in low-resource language pairs.

Our results highlight key factors that contribute to effective direct Persian–English speech-to-speech translation (S2ST) in low-resource conditions. First, self-supervised pretraining of the encoder significantly enhances model performance. The pretrained Conformer encoder captures robust acoustic and phonetic patterns from large amounts of unlabeled Persian speech, enabling the fine-tuned S2ST system to generalize more effectively than models trained from scratch.

Second, discrete acoustic units provide a strong intermediate representation for S2ST. By decoupling linguistic content modeling from waveform synthesis, the encoder–decoder can focus on cross-lingual mappings, while the unit vocoder specializes in producing natural speech. This division of responsibilities allows unit-based S2ST systems to match or surpass spectrogram-based approaches when combined with appropriate pretraining.

%Bibliography
\bibliographystyle{unsrt}  
\bibliography{references}

\end{document}